\documentclass[11pt]{article}

\usepackage[preprint]{acl}

\usepackage{times}
\usepackage{latexsym}
\usepackage[T1]{fontenc}
\usepackage[utf8]{inputenc}
\usepackage{microtype}
\usepackage{inconsolata}

\usepackage{graphicx}
\usepackage{booktabs}
\usepackage{multirow}
\usepackage{amsmath}
\usepackage{amssymb}
\usepackage{mathtools}
\usepackage{amsthm}
\usepackage{enumitem}
\usepackage[most]{tcolorbox}
\usepackage[capitalize,noabbrev]{cleveref}

\theoremstyle{plain}

\theoremstyle{definition}

\theoremstyle{remark}

\title{Decoupled Mixture-of-Experts for Parametric Knowledge Injection}

\usepackage{authblk}
\author[1,*]{\textbf{Baoqing Yue}}
\author[1,*]{\textbf{Weihang Su}}
\author[1]{\textbf{Qingyao Ai}}
\author[1]{\textbf{Yichen Tang}}
\author[1]{\textbf{Changyue Wang}}
\author[1]{\\ \textbf{Jiacheng Kang}}
\author[1]{\textbf{Jingtao Zhan}}
\author[1]{\textbf{Yiqun Liu}}

\affil[1]{Department of Computer Science and Technology, Tsinghua University}
\affil[*]{Equal contribution.}

\begin{document}
\maketitle
\begin{center}
\end{center}
\begin{abstract}

Knowledge injection aims to equip large language models (LLMs) with external, domain-specific, or time-sensitive knowledge. 
Existing approaches typically face a trade-off between flexibility and integration: retrieval-augmented generation keeps knowledge outside the model but only provides prompt-level augmentation, whereas post-training based methods encode new knowledge into shared parameters but may introduce catastrophic forgetting, knowledge conflict, and costly updates. 
In this paper, we propose Decoupled Mixture-of-Experts (DMoE), a modular architecture for parametric knowledge injection that decouples both experts and the router from the base model. 
DMoE converts external knowledge corpora into independently updatable expert modules and uses a lightweight uncertainty-aware router to activate relevant experts only when the base model lacks sufficient knowledge during generation. 
To support efficient auto-regressive inference, DMoE attaches experts only to the final-layer feed-forward network, preserving KV-cache reuse while enabling parameter-level knowledge augmentation. 
Experiments on knowledge-intensive benchmarks show that DMoE consistently improves answer quality over retrieval and adapter-based baselines.

\end{abstract}

\begin{figure}[t]
\includegraphics[width=\columnwidth]{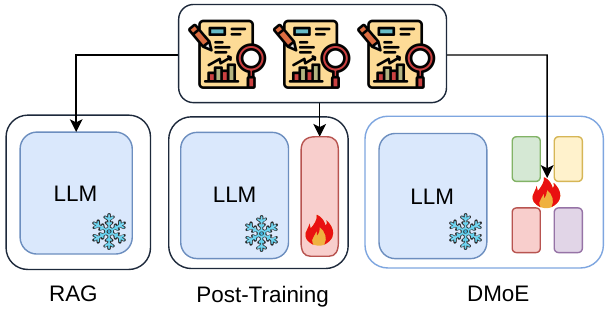}
\caption{Comparison of knowledge injection paradigms. RAG injects knowledge as external context, while post-training modifies shared model parameters and may introduce conflict or forgetting. DMoE decouples knowledge modules from the base model, enabling modular and efficient parametric integration.}
\label{fig:intro}
\end{figure}

\section{Introduction}

Large Language Models (LLMs) have demonstrated strong generalization capabilities across a wide range of tasks \cite{brown2020language,chowdhery2023palm}.
However, their parametric knowledge is inevitably static after pre-training.
As a result, LLMs often fail on domain-specific or time-sensitive queries, producing hallucinated or outdated responses at inference time \cite{song2025injecting,xu2024hallucination}.
This limitation has motivated growing interest in {knowledge injection}, which aims to equip LLMs with external knowledge during inference or through post-training methods \cite{ovadia2023fine,lauscher2020common,ai2025memorybench}.

As illustrated in Figure~\ref{fig:intro}, existing knowledge injection methods can be broadly grouped into retrieval-based and post-training-based paradigms.
Retrieval-Augmented Generation (RAG) keeps knowledge outside the model and dynamically augments the input with retrieved documents \cite{borgeaud2022improving,lewis2020retrieval}.
This design makes knowledge easy to update, since the retrieval corpus can be modified without changing model parameters.
However, the injected knowledge remains at the prompt level: it is only exposed to the model as additional context, rather than being integrated into the model's parameter space.
Consequently, RAG provides flexible but relatively shallow knowledge augmentation, and its inference efficiency can be limited by repeated retrieval and long-context processing.

In contrast, post-training-based methods, including Supervised Fine-Tuning (SFT) \cite{wang2022super,mishra2021cross} and parameter-efficient variants such as LoRA \cite{hu2022lora}, encode new knowledge directly into model parameters.
Although this enables deeper parameter-level integration, the injected knowledge is still written into a shared parameter space that already stores diverse pretrained knowledge.
When knowledge is continuously updated or expanded, such shared updates may interfere with existing capabilities, introduce knowledge conflict, or require repeated re-training as the external corpus changes.
Thus, while post-training-based approaches integrate knowledge more deeply than RAG, they often sacrifice modularity, update efficiency, and knowledge isolation.

This trade-off reveals a deeper architectural bottleneck.
Most existing LLMs organize knowledge either as external prompt context or as entangled dense parameters.
Neither form provides an explicit mechanism for isolating heterogeneous knowledge, routing to relevant knowledge modules, or incrementally expanding the model without perturbing unrelated knowledge.
A desirable knowledge injection architecture should therefore satisfy three requirements: it should integrate knowledge at the parameter level, keep injected knowledge modular and independently updatable, and preserve efficient auto-regressive inference.

To this end, we propose \textbf{Decoupled Mixture-of-Experts (DMoE)}, a modular architecture for parametric knowledge injection.
Inspired by the Mixture-of-Experts (MoE) principle of conditional computation and expert specialization \cite{jacobs1991adaptive,shazeer2017outrageously,fedus2022switch}, DMoE differs from conventional MoE architectures in that both the experts and the router are decoupled from the base model.
Given an external knowledge corpus, DMoE partitions the corpus into knowledge units and constructs lightweight expert modules while keeping the base model unchanged.
These experts are stored outside the dense backbone and can be added, removed, or updated independently.
During inference, a lightweight uncertainty-aware router estimates whether the current query requires external expert support and activates relevant experts only when necessary.

A key design goal of DMoE is to support efficient autoregressive generation.
Naively attaching dynamically selected experts to multiple transformer layers can make cache reuse difficult when the active expert set changes after the prefix has been processed, since the cached key-value states would no longer correspond to the hidden states produced under the newly activated experts.
DMoE avoids this issue by attaching experts after the attention computation of the final transformer layer, specifically to the final-layer feed-forward network.
Because expert activation does not modify the representations from which earlier-layer key-value caches are computed, DMoE can reuse cached attention states during autoregressive generation.
As a result, DMoE enables parameter-level knowledge augmentation while substantially reducing inference overhead compared with dynamic retrieval or multi-layer expert injection strategies.

We evaluate DMoE on a suite of knowledge-intensive benchmarks, focusing on both answer quality and inference efficiency.
Experimental results show that DMoE consistently improves over dense-model baselines and remains competitive with retrieval- and adapter-based knowledge injection methods.
At the same time, DMoE substantially reduces inference overhead by preserving KV-cache reuse.
Further ablation studies validate the main architectural choices: decoupled experts are better suited for knowledge injection than conventional coupled MoE variants, uncertainty-aware routing remains robust across triggering thresholds, and final-layer FFN attachment yields the best effectiveness--efficiency trade-off.

In summary, our contributions are as follows:
\begin{itemize}[leftmargin=*]
\item We propose DMoE, which decouples knowledge experts and the router from the base model, enabling modular and independently updatable knowledge injection.

\item DMoE uses uncertainty-aware routing to selectively activate relevant experts and attaches them to the final-layer feed-forward network, preserving KV-cache reuse during generation.

\item Experimental results show that DMoE improves answer quality over dense baselines while achieving competitive performance with substantially lower inference overhead.
\end{itemize}

\section{Related Work}
\label{sec:related_work}

\subsection{Retrieval-Augmented Generation}
Retrieval-Augmented Generation (RAG) injects external knowledge into LLMs by retrieving relevant information from external repositories and conditioning generation on the retrieved context~\cite{lewis2020retrieval,dong2025decoupling,tu2025robust,su2025dynamic,su2025surge}.
By keeping knowledge outside model parameters, RAG provides a flexible mechanism for improving factual grounding, mitigating hallucinations~\cite{wang2026joint,su2025towards,su2024unsupervised,wang2025decoupling-2}, supporting knowledge updates~\cite{wang2025decoupling,wang2025knowledge}, and adapting LLMs to specialized domains without full model retraining~\cite{su2024stard,su2025judge,su2025pre,su2026enhancing}.
Most RAG systems follow a retrieval-then-read pipeline, where a search module retrieves documents from a large-scale corpus~\cite{robertson2009probabilistic,su2024wikiformer,fang2024scaling} and the generator uses them as additional input context.
Recent extensions further explore dynamic RAG~\cite{jiang2023active,su2024dragin,su2024mitigating}, graph-based RAG~\cite{edge2024local}, parametric RAG~\cite{su2025parametric,su2025sigirap,su2026enhancing,su2026decoupling}, and agentic RAG~\cite{jin2025search,su2026skill}.
Despite these advances, RAG primarily injects knowledge at the prompt level: the retrieved evidence is exposed to the model as external context rather than integrated into its parameter space.
As a result, RAG remains highly updateable but provides relatively shallow knowledge incorporation, and its inference efficiency can be limited by repeated retrieval and long-context processing.

\subsection{Post-training Based Knowledge Injection}
Post-training methods inject knowledge by further optimizing model parameters using data from external sources.
Supervised Fine-Tuning (SFT)~\cite{mishra2021cross,ouyang2022training,taori2023alpaca} commonly trains the model on synthetic or human-annotated instruction--response pairs, enabling deeper parameter-level incorporation of new knowledge than context-only augmentation.
However, directly updating the base model's shared parameters may interfere with previously acquired knowledge, leading to catastrophic forgetting and knowledge conflict~\cite{goodfellow2013empirical,kemker2018measuring}.
To reduce training cost and limit parameter interference, parameter-efficient fine-tuning (PEFT) methods~\cite{houlsby2019parameter,han2024parameter}, such as LoRA~\cite{hu2022lora}, prompt tuning~\cite{lester2021power}, and prefix tuning~\cite{li2021prefix}, freeze most base parameters and train only a small set of additional parameters.
Although PEFT substantially improves update efficiency, standard PEFT methods are typically trained as task- or domain-level adapters and do not by themselves provide fine-grained knowledge isolation, expert-level routing, or independent updates for individual knowledge units.
DMoE builds on the efficiency of lightweight parameter modules while organizing them as decoupled, retrievable experts for modular knowledge injection.

\subsection{Previous Explorations in Decoupling MoE}

Several works have attempted to decouple components of MoE to improve efficiency or training stability. Read-ME \cite{cai2024read} introduces a pre-gating router partially decoupled from the MoE backbone to enable expert-aware batching and caching. EvoMoE \cite{nie2021evomoe} decouples expert training from sparse gating through a progressive dense-to-sparse evolution scheme. StableMoE \cite{dai2022stablemoe} distills and then freezes a router decoupled from the backbone to stabilize routing. DeMo \cite{wang2025decoupled} designs a feature-level decoupled MoE for multi-modal object re-identification, focusing on modality-specific feature weighting. Unlike previous works that decouple either the router or the training phases, our approach \emph{fully} decouples both experts and router from the base model, isolating each unit of knowledge and enabling a scalable routing mechanism.

\section{Methodology}

\begin{figure*}[t]
    \centering
    \includegraphics[width=\textwidth]{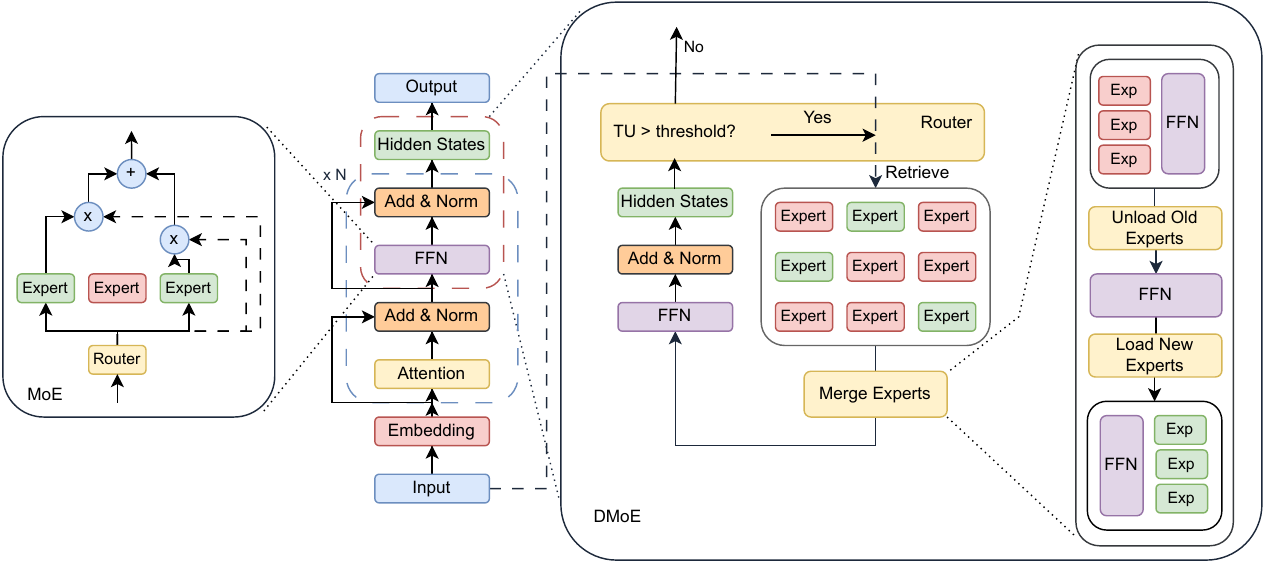}
    \caption{
        \textbf{Comparison of architectures} between the dense model, traditional Mixture-of-Experts (MoE), and the proposed Decoupled Mixture-of-Experts (DMoE).
        In the MoE architecture, the feed-forward layers of a dense model are replaced with a coupled router-expert network.
        In contrast, DMoE decouples both the router and experts from the base model.
        During inference, DMoE retrieves and updates relevant experts when knowledge injection is required, enabling adaptive and non-destructive knowledge integration.
    }
    \label{fig:main}
\end{figure*} 

\subsection{Preliminaries and Task Formulation}

We first characterize the effect of knowledge injection through its impact on the model's predictive uncertainty, following prior uncertainty-aware retrieval and routing methods \cite{jiang2023active,su2024dragin}. At each decoding step $t$, the model produces a softmax distribution $p_t$ over the vocabulary. We define the token uncertainty (TU) at step $t$ as the entropy of this distribution:
\begin{equation}
    \mathrm{TU}_t = -\sum_{v} p_t(v)\log p_t(v).
\end{equation}

\noindent During auto-regressive generation, the model does not have access to the correct next token and thus cannot directly assess whether its prediction is accurate or whether additional knowledge is required. TU therefore serves as a practical inference-time signal that reflects how uncertain the model is about its next-token distribution: a larger TU indicates that the model lacks sufficient internal knowledge for the current input and is more likely to benefit from external knowledge injection.

Formally, let $\mathcal{K} = \{\mathcal{K}_1, \mathcal{K}_2, \dots, \mathcal{K}_N\}$ denote a set of external knowledge units, each representing a semantically coherent piece of information (e.g., a fact, a document, or a reasoning pattern). The objective of knowledge injection is to construct an augmented model $\theta'$ by incorporating a subset of knowledge $\mathcal{K}^* \subseteq \mathcal{K}$, so as to reduce TU on knowledge-relevant samples $\mathcal{D}_{\mathrm{know}}$:
\begin{equation}
    \theta' = \mathcal{A}(\theta, \mathcal{K}^*), 
\end{equation}
\begin{equation}
    \mathcal{K}^* = \arg\min_{\mathcal{S} \subseteq \mathcal{K}} 
    \mathbb{E}_{x \sim \mathcal{D}_{\mathrm{know}}}
    \left[ \mathrm{TU}(x; \mathcal{A}(\theta, \mathcal{S})) \right],
\end{equation}
where $\mathcal{A}(\cdot)$ denotes a knowledge augmentation mechanism.
Existing approaches differ in how they realize this objective. In RAG settings, the model parameters remain unchanged, and external knowledge $\mathcal{K}^*$ is injected through the input context, resulting in a shallow form of knowledge injection. 
In contrast, parameter-level knowledge injection methods such as post-training modify $\theta$ directly to encode $\mathcal{K}^*$, enabling the model to internalize new knowledge.



\subsection{Model Architecture}

As illustrated in Figure~\ref{fig:main}, \textbf{DMoE} architecture consists of three main components: a frozen \textit{base model}, a \textit{router}, and a large set of \textit{experts}. Unlike conventional MoE architectures that replace the feed-forward network (FFN) layers of a dense model with a jointly trained gating network and multiple experts, DMoE decouples both router and experts from the base model.

\subsubsection{Experts}
\begin{figure}[t]
    \centering
    \includegraphics[width=\linewidth]{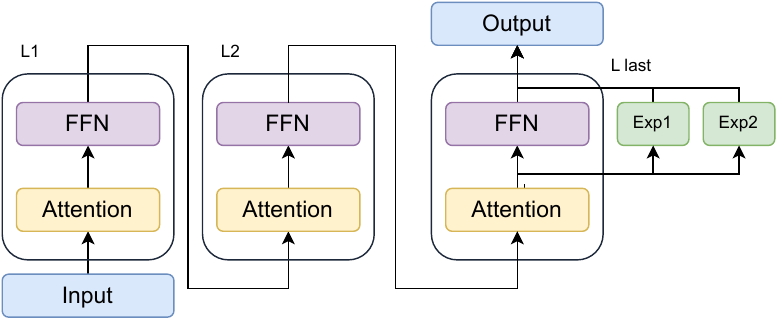}
    \caption{
        \textbf{Illustration of expert placement in DMoE.}
        Experts are attached only to the FFN of the final Transformer layer so that the hidden states feeding earlier attention blocks remain unchanged. 
        Placing experts at intermediate layers would modify representations consumed by subsequent attention modules, preventing efficient reuse of KV-cache during decoding.
    }
    \label{fig:last-layer-expert}
\end{figure}

\paragraph{Expert architecture.}
In DMoE, each expert represents a semantic unit of knowledge, which is practically realized via lightweight adapters (e.g., LoRA~\cite{hu2022lora} or other PEFT variants) attached to the FFN layers, which are known to store factual and semantic knowledge~\cite{geva2021transformer,nanda2023fact,yu2024neuron}. 
To maintain practical KV-cache reuse during decoding, we attach experts only to the final-layer FFN (Figure~\ref{fig:last-layer-expert}); a detailed analysis of this design choice is provided in Section~\ref{sec:ablation_expert_layer}.

\paragraph{Expert construction.}
For each knowledge unit $\mathcal{K}_i$, we construct an expert by training a lightweight parameter update $\Delta\theta_i$ on data derived from that unit.
Following PRAG~\cite{su2025parametric}, we first augment each document into instruction-style training instances and then optimize a PEFT module while keeping the base model frozen.
This produces a collection of knowledge-specific experts $\{\Delta\theta_i\}_{i=1}^{N}$, where each expert is associated with both a parameter update and a text surrogate used for routing.
In this work, we use this construction pipeline as a transparent instantiation of DMoE, while our main focus is on how independently trained experts are organized, routed, and integrated during inference.

\subsubsection{Router}                                          

\subparagraph{Triggering.}  
Routing is activated only when the current model exhibits insufficient knowledge of the input.
Since our goal is to reduce TU, there is no need to involve additional experts when the model is already confident.
At each decoding step $t$, we trigger routing only when TU is above a given threshold:
\begin{equation}
    \text{Trigger}_t = \mathbb{I}\big[\mathrm{TU}_t > \tau\big],
\end{equation}
where $\tau$ is a given threshold.
When $\text{Trigger}_t = 1$, the router deactivates outdated experts and activates relevant experts; otherwise, decoding proceeds with the current model.

\paragraph{Routing.}
To make knowledge injection both effective and scalable, the router should (i) select experts that most likely reduce token uncertainty (TU) under the current context, while (ii) remaining fully \emph{decoupled} from the base model so that experts can be added/removed/updated without retraining any neural router.
We therefore adopt a \emph{BM25-based lexical router} implemented via an external inverted index.

Each expert $E_i$ is trained on a knowledge unit $\mathcal{K}_i$ and is associated with a text surrogate $\mathcal{D}_i$ (the document plus its augmentations used in expert training; see Section~4.1.3).
At decoding step $t$, we form a routing query $q_t$ from the task input and the current generation context, and retrieve experts whose associated texts best match $q_t$ under BM25:
When a high-TU position triggers routing, $q_t$ is constructed from the generated prefix up to, but excluding, that position, so the retrieved experts are conditioned on exactly the context already represented in the KV cache.
\begin{equation}
\label{eq:bm25_routing}
E_{\text{sel}} \;=\; \operatorname{Top\mbox{-}k}_{i \in [N]} \;\mathrm{BM25}(q_t, \mathcal{D}_i).
\end{equation}
This design is training-free, incrementally updatable (adding an expert only inserts $\mathcal{D}_i$ into the index), and avoids introducing an additional neural encoder that would partially re-couple routing with model parameters.
Finally, we activate the selected experts by composing them with the frozen base model only as effective parameters for the current decoding state:
\begin{equation}
\label{eq:expert_merge}
\theta^{\mathrm{eff}}_t
\;=\;
\theta
\;+\;
\sum_{E_i \in E_{\text{sel}}} \Delta\theta_i .
\end{equation}
Here, $\theta$ denotes the unchanged base-model parameters, while $\theta^{\mathrm{eff}}_t$ denotes the temporary parameter composition used for decoding under the currently active experts.
Generation resumes from the triggering token, and this active expert set is kept until the next triggering event, avoiding redundant expert swaps at every decoding step.

\begin{table*}[t]
  \caption{
    \textbf{Main results on four knowledge-intensive benchmarks.}
    We report EM/F1 or ACC across datasets and base models. Best values are in bold.
  }
  \label{tab:main_results}
  \begin{center}
    \begin{small}
      \begin{sc}
        \setlength{\tabcolsep}{6pt}
        \begin{tabular}{l|cc|cc|cc|c}
          \toprule
          \multirow{2}{*}{\textbf{Method}} &
          \multicolumn{2}{c|}{\textbf{CWQ}} &
          \multicolumn{2}{c|}{\textbf{HotpotQA}} &
          \multicolumn{2}{c|}{\textbf{Quasar-T}} &
          \textbf{StrategyQA} \\
          \cmidrule(lr){2-8}
          & \textbf{EM} & \textbf{F1}
          & \textbf{EM} & \textbf{F1}
          & \textbf{EM} & \textbf{F1}
          & \textbf{ACC} \\
          \midrule
          \multicolumn{8}{l}{\textbf{Base Model: Llama3.2-1B}} \\
          \midrule
          Basic-RAG   & 0.1633 & 0.2384 & 0.1700 & 0.2463 & 0.2800 & 0.3513 & 0.4333 \\
          FLARE       & 0.2400 & 0.3154 & 0.0733 & 0.1303 & 0.1867 & 0.2190 & 0.5367 \\
          PRAG        & \textbf{0.2500} & 0.3284 & 0.0733 & 0.1427 & 0.2200 & 0.2514 & 0.5600 \\
          SFT-LoRA    & 0.2167 & 0.3092 & 0.0767 & 0.1325 & 0.2133 & 0.2481 & 0.5533 \\
          DMoE (Ours) & 0.2467 & \textbf{0.3479} & \textbf{0.1800} & \textbf{0.2553} & \textbf{0.3133} & \textbf{0.3658} & \textbf{0.5667} \\
          \midrule
          \multicolumn{8}{l}{\textbf{Base Model: Qwen2.5-1.5B}} \\
          \midrule
          Basic-RAG   & 0.1233 & 0.1888 & \textbf{0.1567} & \textbf{0.2446} & 0.1833 & 0.2549 & 0.3667 \\
          FLARE       & 0.1200 & 0.1806 & 0.1000 & 0.1717 & 0.1733 & 0.2242 & 0.2500 \\
          PRAG        & \textbf{0.2167} & 0.3050 & 0.1033 & 0.1583 & 0.1733 & 0.2312 & 0.5833 \\
          SFT-LoRA    & 0.1967 & 0.3092 & 0.1033 & 0.1595 & 0.1767 & 0.2292 & 0.5733 \\
          DMoE (Ours) & \textbf{0.2167} & \textbf{0.3285} & 0.1533 & 0.2375 & \textbf{0.2333} & \textbf{0.2795} & \textbf{0.6133} \\
          \bottomrule
        \end{tabular}
      \end{sc}
    \end{small}
  \end{center}
  \vskip -0.1in
\end{table*}

\section{Experimental Setup}

\subsection{Benchmarks and Metrics}
We conduct comprehensive evaluations on a diverse suite of knowledge-intensive benchmarks, each representing a distinct aspect of the knowledge injection problem.  
\textbf{HotpotQA} \citep{yang2018hotpotqa} focuses on multi-hop reasoning, where the model must integrate evidence from multiple documents to answer complex questions that require reasoning across facts.  
\textbf{ComplexWebQuestions} \citep{talmor2018web} extends this setting to an open-domain environment, challenging the model to perform compositional reasoning and inject relevant knowledge from large-scale web content.  
\textbf{Quasar-T} \citep{dhingra2017quasar} further evaluates open-domain retrieval and reasoning by requiring models to locate and synthesize factual knowledge from a large corpus to answer trivia-style questions.  
\textbf{StrategyQA} \citep{geva2021did} assesses implicit multi-hop reasoning, where the required knowledge is not explicitly stated in the question and must be inferred or retrieved based on world knowledge.  

For evaluation, we primarily assess the effectiveness of knowledge injection methods in the main results. 
Effectiveness is measured using EM and F1 (ACC for StrategyQA), which reflect how well a model integrates and utilizes injected knowledge. 
We analyze efficiency separately in Section~\ref{sec:efficiency_analysis} using the average inference time and average GPU memory consumption per sample.

\subsection{Baselines}

We compare DMoE against representative baselines spanning the major paradigms of knowledge injection: Basic-RAG, FLARE, PRAG, and SFT-LoRA, all built on \textit{Llama-3.2-1B-Instruct} and \textit{Qwen2.5-1.5B-Instruct}. For an equitable comparison, all retrieval-based methods share the same retrieval corpus of 27{,}613 passage-level documents, the same retriever (Elasticsearch BM25), retrieval budget ($k=3$), and greedy decoding; retrieval queries are truncated to 128 tokens, and the corpus is not additionally chunked because each item is already a short single-topic passage of approximately 100 words. Basic-RAG performs one-shot retrieval and appends the retrieved passages to the prompt, FLARE retrieves during decoding and therefore cannot reuse the KV-cache, PRAG combines retrieved passages with a corresponding parametric adapter using the same augmentation-and-adapter recipe as our expert construction pipeline, and SFT-LoRA trains a single LoRA adapter on supervision merged across the full corpus.

\subsection{Implementation Details}

Before training experts, each base model undergoes a one-epoch held-out format-stabilization warmup.
Concretely, warmup uses the same task and system instructions as the main evaluation, but only evaluation-disjoint instances, and is intended to stabilize output formatting consistency rather than inject factual or benchmark-specific knowledge.
This stage is retained to match the PRAG recipe and maintain a controlled comparison; its impact is analyzed in Section~\ref{sec:ablation_warmup}.

The external knowledge corpus used to construct experts is drawn from the Wikipedia corpus used in DPR \cite{karpukhin2020dense}. 
This pool is used by all retrieval-based baselines and by DMoE, so comparisons are controlled under the same accessible knowledge source.
Importantly, corpus construction and expert training do not use answer strings, gold supporting-document annotations, or any manually labeled rationales.
Each document is augmented only from its own passage text by: (1) a single paraphrased rewrite that preserves its factual content, and (2) three question-answer pairs generated from the document.
These augmentations enrich the supervision signals for expert training and help each expert better capture the underlying knowledge unit.
For each knowledge unit, we train a dedicated expert implemented as a LoRA adapter with rank~4, scaling factor $\alpha=16$, and learning rate $1\times 10^{-5}$. 
Experts are trained for one epoch and inserted only into the final-layer feed-forward network to ensure full compatibility with KV-cache reuse during autoregressive decoding.
During inference, we set triggering threshold to $2.0$, use greedy decoding and activate the top-$k$ experts ($k=3$ by default). 
We use BM25 to construct expert and context representations. All efficiency measurements (inference time and GPU memory usage) are conducted on a single NVIDIA A100 80GB GPU.

\section{Experimental Results}

\subsection{Overall Results}

Table~\ref{tab:main_results} summarizes overall effectiveness across all baselines.  We set test corpus size to 300 for all benchmarks.
Table~\ref{tab:main_results} shows a general advantage for DMoE. 
Across the 14 reported effectiveness metrics, DMoE obtains the best or tied-best score on 11 metrics, including the strongest F1 on CWQ for both base models, both Quasar-T metrics for both base models, and StrategyQA ACC for both base models. 
This pattern suggests that DMoE is usually the strongest method in the reported comparison.
The exceptions are concentrated rather than scattered. 
On CWQ, PRAG slightly exceeds DMoE in EM for Llama3.2-1B and ties it for Qwen2.5-1.5B, although DMoE has the higher CWQ F1 in both cases. 
On HotpotQA with Qwen2.5-1.5B, Basic-RAG is marginally better than DMoE on both EM and F1, while DMoE is best on both HotpotQA metrics with Llama3.2-1B. 
Thus, Table~\ref{tab:main_results} supports a measured conclusion: DMoE is broadly competitive and most often best among the evaluated methods, but the results do not establish dominance on every dataset, base model, or metric.

\subsection{Efficiency Analysis}
\label{sec:efficiency_analysis}

\begin{table}[t]
  \caption{
    \textbf{Efficiency analysis.}
    We report all available average inference latency and GPU memory measurements per sample. Lower is better, and best values within each base-model block are in bold.
  }
  \label{tab:efficiency_results}
  \begin{center}
    \begin{small}
      \begin{sc}
        \setlength{\tabcolsep}{4pt}
        \begin{tabular}{lcc}
          \toprule
          \textbf{Method} & \textbf{Time (s)} & \textbf{GPU (GB)} \\
          \midrule
          Basic-RAG   & 1.8900 & \textbf{2.5400} \\
          FLARE       & 9.2643 & 13.9718 \\
          PRAG        & \textbf{1.3600} & 4.8300 \\
          SFT-LoRA    & 1.6700 & 4.8200 \\
          DMoE (Ours) & 2.6656 & 7.2352 \\
          \bottomrule
        \end{tabular}
      \end{sc}
    \end{small}
  \end{center}
  \vskip -0.1in
\end{table}

As shown in Table~\ref{tab:efficiency_results}, static baselines such as Basic-RAG, PRAG, and SFT-LoRA can incur lower per-sample cost because they do not modify retrieval or routing decisions during decoding. Among methods that adapt generation online, however, DMoE is roughly $3\times$ faster than FLARE while also using substantially less memory. The core reason is KV-cache reuse. FLARE-style dynamic RAG interleaves generation with retrieval and prompt rewriting, which grows the context over time and invalidates cache reuse; consequently, attention over an expanding prefix is repeatedly recomputed, inflating both latency and memory.

In contrast, DMoE keeps the prompt state stable and performs adaptation by conditionally activating lightweight final-layer experts. Since experts are confined to the last FFN layer, the KV-cache remains fully valid throughout decoding, largely avoiding the context-growth penalty and redundant attention computation. This cache-safe design translates into consistent time and memory advantages in practice: DMoE reduces GPU memory usage by roughly $1.6\times$--$1.9\times$ compared to FLARE.

\begin{figure}[t]
    \centering
    \includegraphics[width=1.0\linewidth]{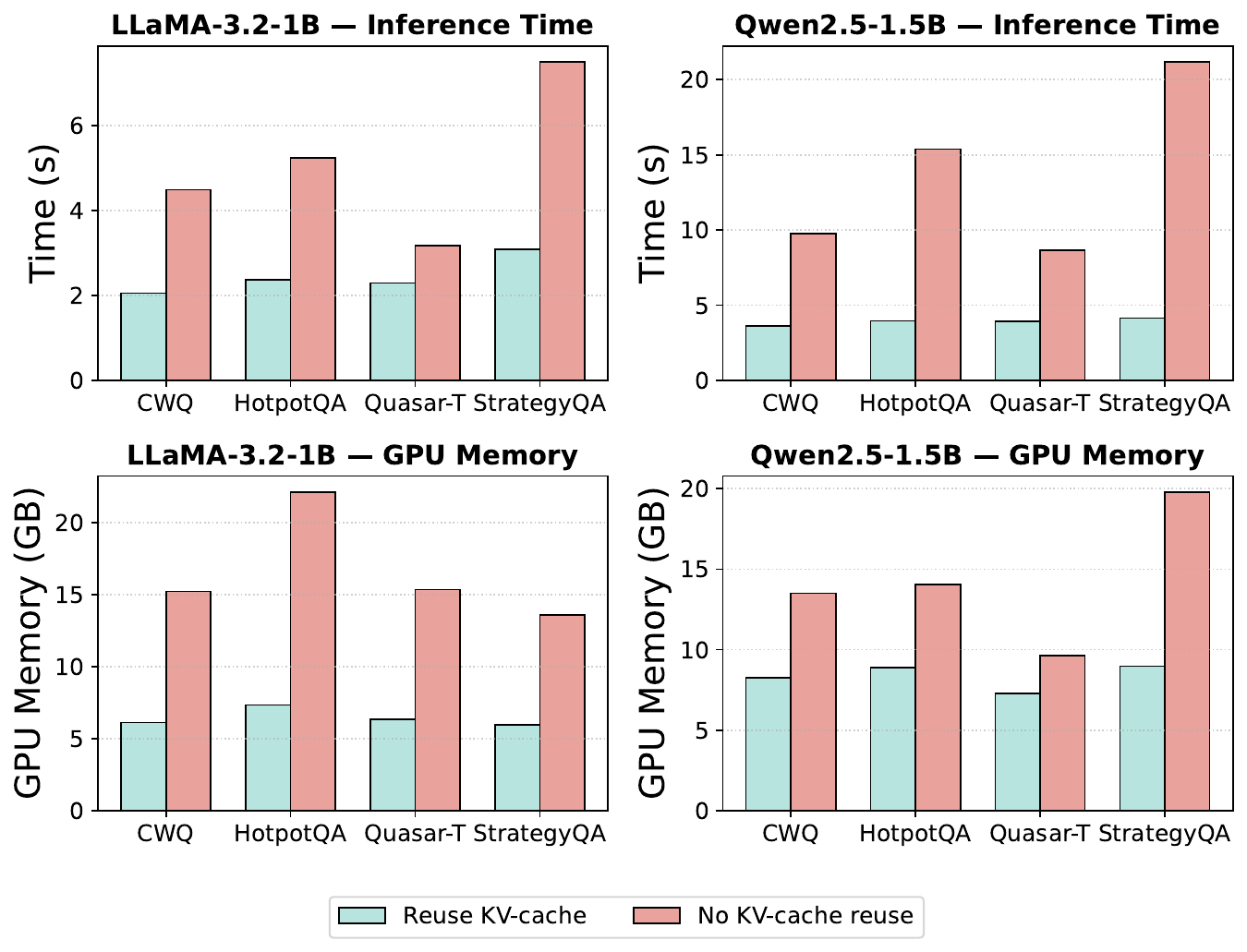}
    \caption{
    \textbf{KV-cache reuse substantially improves autoregressive inference efficiency.}
    We report the inference latency (top) and GPU memory footprint (bottom) for two base models (Llama-3.2-1B and Qwen2.5-1.5B) across four benchmarks, comparing \textit{Reuse KV-cache} vs.\ \textit{No KV-cache reuse}.
    KV-cache reuse consistently reduces both time and memory across all datasets.
}

    \label{fig:kv-cache-speedup}
\end{figure}

Figure~\ref{fig:kv-cache-speedup} further confirms the centrality of KV caching: across all benchmarks and both base models, enabling KV caching yields a $1.3\times$--$5.1\times$ latency speedup and a $1.2\times$--$2.5\times$ reduction in GPU memory usage, motivating cache-compatible expert placement.

\setcounter{table}{3}
\begin{table*}[t]
  \caption{
  \textbf{DMoE vs.\ a coupled MoE backbone.}
  We compare DMoE against a finetuned MoE model (OLMoE-1B-7B) under SFT-LoRA.
  DMoE achieves higher effectiveness (EM/F1/ACC) while using markedly lower inference time and GPU memory.
  Higher is better for EM/F1/ACC; lower is better for Time/GPU. Best values are in bold.}
  \label{tab:dmoe_vs_moe}
  \begin{center}
    \begin{small}
      \begin{sc}
        \setlength{\tabcolsep}{2.5pt}
        \resizebox{\textwidth}{!}{%
        \begin{tabular}{l|cc|cc|cc|c|cc}
          \toprule
          \multirow{2}{*}{\textbf{Method}} &
          \multicolumn{2}{c|}{\textbf{CWQ}} &
          \multicolumn{2}{c|}{\textbf{HotpotQA}} &
          \multicolumn{2}{c|}{\textbf{Quasar-T}} &
          \textbf{StrategyQA} &
          \multicolumn{2}{c}{\textbf{Efficiency}} \\
          \cmidrule(lr){2-10}
          & \textbf{EM} & \textbf{F1} & \textbf{EM} & \textbf{F1} & \textbf{EM} & \textbf{F1} & \textbf{ACC} & \textbf{Time (s)} & \textbf{GPU (GB)} \\
          \midrule
          DMoE (Llama3.2-1B)  & \textbf{0.2467} & \textbf{0.3479} & \textbf{0.1800} & \textbf{0.2553} & \textbf{0.3133} & \textbf{0.3658} & 0.5667 & \textbf{2.6656} & \textbf{7.2352} \\
          DMoE (Qwen2.5-1.5B) & 0.2167 & 0.3285 & 0.1533 & 0.2375 & 0.2333 & 0.2795 & \textbf{0.6133} & 3.9138 & 8.3415 \\
          MoE (SFT-LoRA, OLMoE-1B-7B) & 0.1500 & 0.2638 & 0.1033 & 0.1790 & 0.2100 & 0.2579 & 0.5900 & 20.0217 & 26.0843 \\
          \bottomrule
        \end{tabular}%
        }
      \end{sc}
    \end{small}
  \end{center}
  \vskip -0.1in
\end{table*}
\setcounter{table}{2}

\begin{table}[t]
  \caption{Performance (EM) when reusing KV-cache while introducing experts into different portions of FFN layers. Here, $x\%$ denotes inserting experts into the final $x\%$ of FFN layers, while \textsc{Only-last} inserts experts only into the final-layer FFN. Only-last preserves KV-cache compatibility and achieves the best overall accuracy across datasets.}

  \label{tab:layer_abl_em_two_models}
  \begin{center}
    \begin{small}
      \begin{sc}
        \setlength{\tabcolsep}{6pt}
        \renewcommand{\arraystretch}{1.15}
        \resizebox{\columnwidth}{!}{%
          \begin{tabular}{lcccc}
            \toprule
            \textbf{Setting} & \textbf{CWQ} & \textbf{HotpotQA} & \textbf{Quasar-T} & \textbf{StrategyQA} \\
            \midrule
            \multicolumn{5}{l}{\textbf{Llama3.2-1B}} \\
            25\%       & 0.2167 & 0.1600 & 0.2967 & 0.5200 \\
            50\%       & 0.2433 & 0.1567 & 0.3067 & 0.5333 \\
            75\%       & 0.2100 & 0.1767 & 0.3100 & 0.5233 \\
            100\%      & 0.2133 & 0.1633 & 0.3000 & 0.5533 \\
            Only-last  & \textbf{0.2467} & \textbf{0.1800} & \textbf{0.3133} & \textbf{0.5667} \\
            \midrule
            \multicolumn{5}{l}{\textbf{Qwen2.5-1.5B}} \\
            25\%       & 0.2033 & 0.1367 & 0.2200 & 0.5633 \\
            50\%       & 0.1767 & 0.1500 & \textbf{0.2400} & 0.5467 \\
            75\%       & 0.1800 & 0.1333 & 0.2233 & 0.5400 \\
            100\%      & 0.1800 & 0.1333 & 0.2200 & 0.5400 \\
            Only-last  & \textbf{0.2167} & \textbf{0.1533} & 0.2333 & \textbf{0.6133} \\
            \bottomrule
          \end{tabular}%
        }
      \end{sc}
    \end{small}
  \end{center}
  \vskip -0.1in
\end{table}

\setcounter{table}{4}

\subsection{Ablation Study}

To better understand the contribution of individual architectural components in DMoE, the main paper focuses on two architectural questions: (1) where experts should be placed to preserve KV-cache compatibility, and (2) whether DMoE's gains can be explained by simply using a conventional MoE backbone.
Additional ablations, including warmup, sensitivity to router top-$k$, TU-triggered routing analyses, LoRA hyperparameter sensitivity, and retriever choice, are reported in Appendix~\ref{sec:appendix_extra_ablation}.

\subsubsection{Impact of Expert Layer Placement}
\label{sec:ablation_expert_layer}

A practical requirement for DMoE is that expert activation must preserve KV-cache reuse during auto-regressive decoding. In a standard transformer decoder, each layer $\ell$ computes
\begin{equation}
    h_{\ell+1} = \mathcal{F}_\ell(h_\ell),
\end{equation}

\noindent where $h_\ell$ is the hidden state, and the keys and values for layer $\ell$'s attention are linear projections of $h_\ell$.
Introducing an expert at any intermediate layer $k$ modifies the hidden state to
\begin{equation}
    \tilde{h}_{k+1} = h_{k+1} + \Delta(h_k),
\end{equation}
which propagates forward to all subsequent layers $j>k$.  
Because the KV-cache stores $K_j^{\text{cached}} = W_K h_j$ and $V_j^{\text{cached}} = W_V h_j$ computed from the unmodified trajectory, any modification \(\tilde{h}_j \neq h_j\) requires recomputing
\begin{equation}
    K_j^{\text{new}} = W_K \tilde{h}_j,\qquad 
V_j^{\text{new}} = W_V \tilde{h}_j,
\end{equation}
breaking KV-cache compatibility and causing substantial latency overhead.

This constraint is avoided only when experts are applied at the \textbf{final FFN layer}.
In a decoder-only model with $L$ layers, the output of the last FFN ($h_{L+1}$) is projected directly to logits and does not feed into any further attention computation.  
Thus, placing experts exclusively at $k=L$ guarantees
\begin{equation}
    \forall j \le L,\quad \tilde{h}_j = h_j,
\end{equation}
ensuring that all cached keys and values remain valid.

Table ~\ref{tab:layer_abl_em_two_models} empirically confirms this analysis.
When experts are inserted into earlier or multiple layers, performance drops sharply due to KV-cache mismatch.
In contrast, attaching experts only to the last FFN layer is generally competitive and often best.
These findings demonstrate a key property of DMoE:  
minimal-intrusive expert placement at the final layer preserves fast decoding while still enabling effective knowledge injection.  
Together with adaptive triggering, this design provides an effective and efficient mechanism for integrating new knowledge during inference.

\subsubsection{DMoE vs.\ Conventional MoE Backbones}

To clarify whether our gains are simply an artifact of adopting an MoE-style backbone, we include a coupled MoE model as a structural control. Specifically, we evaluate \textit{allenai/OLMoE-1B-7B-0125-Instruct} and adapt it with the same SFT-LoRA recipe used in our dense baselines. We choose it because its activated parameter scale is comparable to DMoE, i.e., both systems perform sparse conditional computation with a similar amount of per-token active capacity. This design makes the comparison informative: it helps factor out the confounding effect of increasing total parameters, and tests whether the improvements come from DMoE's architectural decoupling rather than parameter count.
Table~\ref{tab:dmoe_vs_moe} shows that DMoE consistently outperforms the coupled MoE backbone on effectiveness while being substantially more efficient. On CWQ, HotpotQA, and Quasar-T, both DMoE variants achieve higher EM/F1 than OLMoE under SFT-LoRA, and DMoE also matches or exceeds OLMoE on StrategyQA accuracy. More importantly, the system costs differ dramatically: DMoE reduces average inference time from 20.0\,s (OLMoE) to 2.7--3.9\,s, and cuts GPU memory from 26.1\,GB to 7.2--8.3\,GB.

These results support our main design rationale for knowledge injection. A conventional MoE backbone offers conditional computation, but its experts are heavy and must remain resident during inference, and its router/expert parameters are coupled to the base model, which makes knowledge updates harder to localize and incurs high runtime footprint. In contrast, DMoE treats experts as external, lightweight knowledge modules and invokes them only when needed, while keeping the base model intact. Therefore, DMoE's gains cannot be attributed to merely increasing model capacity; they arise from a decoupled, cache-safe modularization that yields a strictly better effectiveness-efficiency trade-off than simply switching to a standard MoE architecture.

\section{Conclusion}

We presented DMoE, a decoupled mixture-of-experts architecture for modular parametric knowledge injection. Unlike conventional retrieval or post-training based methods, DMoE keeps both knowledge experts and the router external to the frozen base model, allowing knowledge units to be added, removed, or updated independently. During inference, DMoE uses uncertainty-aware routing to selectively activate relevant experts and attaches them only to the final-layer feed-forward network, preserving KV-cache reuse during autoregressive generation. Experiments on knowledge-intensive benchmarks show that DMoE consistently improves over dense baselines and achieves competitive performance against retrieval- and adapter-based methods, while substantially reducing the overhead of dynamic knowledge augmentation. These results suggest that decoupled, cache-compatible expert modularization is a promising direction for scalable and updateable knowledge injection in LLMs.



\bibliography{custom}

\clearpage
\appendix

\twocolumn

\section{Additional Ablation Studies}
\label{sec:appendix_extra_ablation}

\subsection{Warmup: Definition and Impact}
\label{sec:ablation_warmup}

Since the focus of this work is the DMoE architecture itself, we do not include dedicated instruction-following benchmarks such as IFEval \citep{zhou2023instruction} to directly measure the effect of warmup on instruction following.
Instead, we assess its impact indirectly through downstream performance on four QA benchmarks.
Warmup yields modest gains of about $+1.0$ EM on CWQ, $+0.3$ EM on HotpotQA, and $+1.3$ EM on Quasar-T, but slightly lowers StrategyQA by about $1.0$ ACC.
This mixed pattern suggests that warmup does not provide a uniform performance benefit and does not indicate a broad degradation in instruction following.
The main setting retains warmup to match the PRAG recipe and maintain a controlled comparison.

\subsection{Router: Impact of Top-$k$ Experts}
\label{sec:appendix_topk}

\begin{figure}[t]
    \centering
    \includegraphics[width=0.82\linewidth]{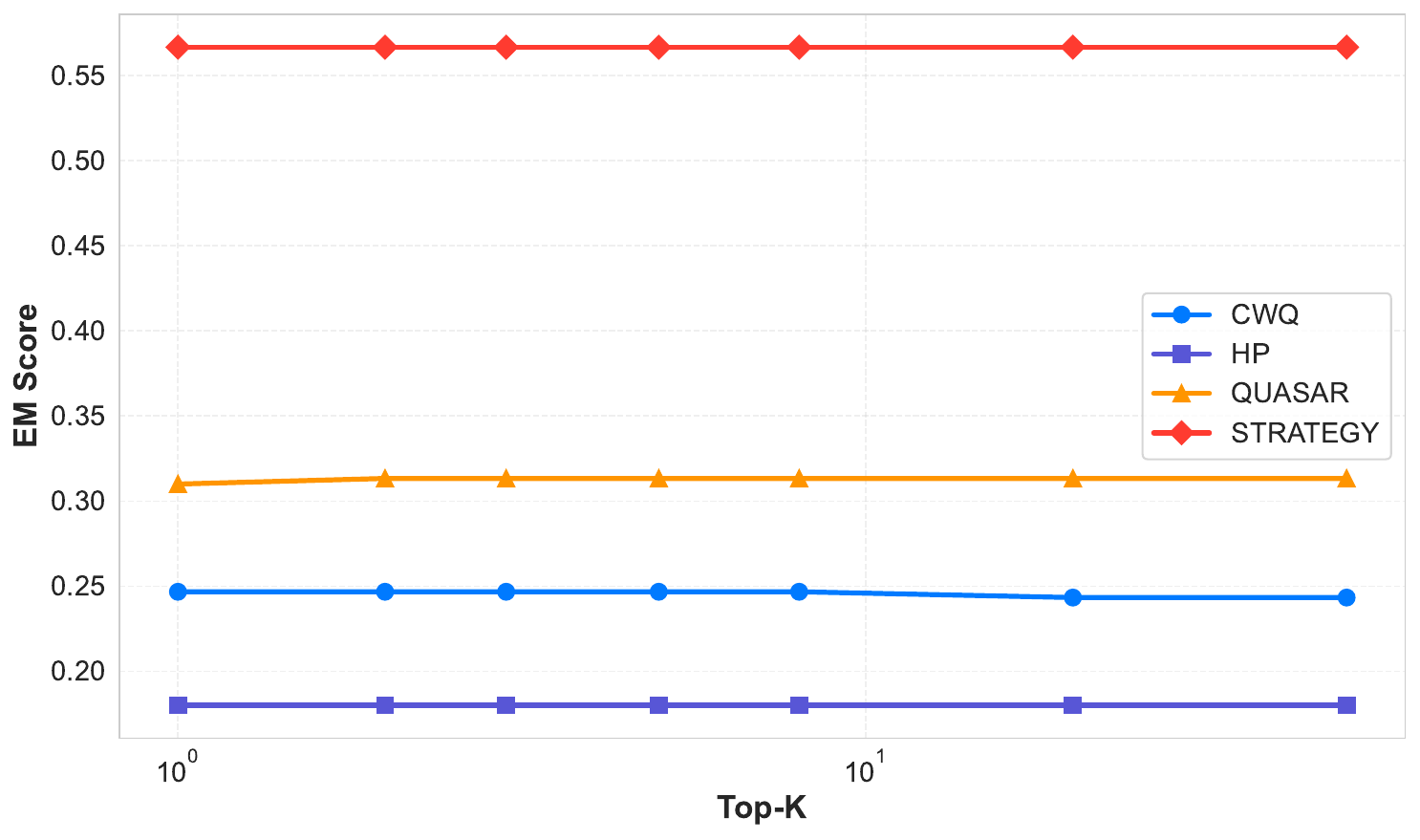}
    \caption{
        \textbf{Impact of the number of activated experts.}
        Varying top-$k$ across the tested range yields nearly unchanged downstream performance on all four benchmarks, indicating that DMoE is robust to this routing hyperparameter.
    }
    \label{fig:topk_abl}
\end{figure}

Figure~\ref{fig:topk_abl} studies sensitivity to the number of activated experts while keeping the remaining setup fixed. The curves remain nearly flat across CWQ, HotpotQA, Quasar-T, and StrategyQA, showing that DMoE is highly robust to the choice of top-$k$. This behavior suggests that the router identifies a sufficiently relevant expert set even with a small activation budget, while increasing the number of activated experts brings only marginal changes in downstream quality. Therefore, the default choice $k=3$ provides a simple and stable operating point without requiring careful tuning.

\subsection{Router: TU-Triggered Routing Analysis}
\label{sec:appendix_tu_trigger_analysis}

This section analyzes token uncertainty (TU) as the trigger signal used by the router. We focus on three questions: whether TU is aligned with downstream performance, how the trigger and router components contribute separately, and how sensitive DMoE is to the triggering threshold.

\subsubsection{TU and Downstream Performance}

\begin{table}[t]
  \caption{\textbf{DMoE accuracy across token-uncertainty bins.} We group 300 evaluation examples into five entropy bins using the base model's token uncertainty (TU) before expert activation and report DMoE performance within each bin. Higher TU consistently corresponds to lower EM/F1, supporting TU as a useful---though imperfect---proxy for when the model lacks sufficient prior knowledge.}
  \label{tab:appendix_tu_bins}
  \begin{center}
    \begin{small}
      \begin{sc}
        \setlength{\tabcolsep}{6pt}
        \renewcommand{\arraystretch}{1.15}
        \begin{tabular}{lccc}
          \toprule
          \textbf{Bin} & \textbf{Entropy range} & \textbf{EM} & \textbf{F1} \\
          \midrule
          1 (low)  & [0.593, 0.863] & 0.4167 & 0.5593 \\
          2        & [0.863, 1.015] & 0.2833 & 0.4213 \\
          3        & [1.017, 1.185] & 0.2333 & 0.3186 \\
          4        & [1.188, 1.365] & 0.1667 & 0.2320 \\
          5 (high) & [1.366, 2.824] & 0.1333 & 0.2085 \\
          \bottomrule
        \end{tabular}
      \end{sc}
    \end{small}
  \end{center}
  \vskip -0.1in
\end{table}

Table~\ref{tab:appendix_tu_bins} sorts 300 evaluation examples by the base model's token uncertainty before expert activation. Low-entropy bins correspond to cases where the base model is already relatively confident, whereas high-entropy bins correspond to cases where the model has more limited prior knowledge. The degradation is monotonic: EM drops from 0.4167 in the lowest-entropy bin to 0.1333 in the highest-entropy bin, and F1 drops from 0.5593 to 0.2085. This trend indicates that TU is aligned with knowledge difficulty and is therefore a reasonable signal for deciding when to trigger expert assistance.

\subsubsection{Component Ablation: Trigger and Router}

\begin{table}[t]
  \caption{\textbf{Component ablation of trigger and router choices.} We disentangle when to activate experts from which experts to select. The full TU-triggered BM25 router performs best, showing that DMoE's gains are not explained by indiscriminate expert activation.}
  \label{tab:appendix_trigger_router_component}
  \begin{center}
    \begin{small}
      \begin{sc}
        \setlength{\tabcolsep}{5pt}
        \renewcommand{\arraystretch}{1.15}
        \resizebox{\columnwidth}{!}{%
        \begin{tabular}{lcc}
          \toprule
          \textbf{Setting} & \textbf{EM} & \textbf{F1} \\
          \midrule
          A. Full DMoE (TU + BM25) & \textbf{0.1800} & \textbf{0.2553} \\
          B. Random trigger + BM25 & 0.1667 & 0.2464 \\
          C. TU + random router & 0.1767 & 0.2406 \\
          D. Always trigger + BM25 & 0.1200 & 0.1665 \\
          E. TU + oracle-style expert router & 0.1767 & 0.2532 \\
          \bottomrule
        \end{tabular}%
        }
      \end{sc}
    \end{small}
  \end{center}
  \vskip -0.1in
\end{table}

Table~\ref{tab:appendix_trigger_router_component} further disentangles the triggering decision from expert selection. The full setting, which combines TU-based triggering with the BM25 router, achieves the best EM and F1. Replacing TU with a random trigger hurts performance, keeping TU but replacing BM25 with a random router also degrades F1, and always triggering experts collapses substantially. These results show that the gains are not driven by indiscriminate expert activation: both deciding when to activate experts and selecting which experts to activate matter. Finally, the oracle-style expert router is only comparable to BM25, suggesting that BM25 is already a strong practical router in this parametric-expert pipeline rather than the dominant bottleneck.

\subsubsection{Impact of Triggering Thresholds}

\begin{figure*}[t]
    \centering
    \includegraphics[width=\textwidth]{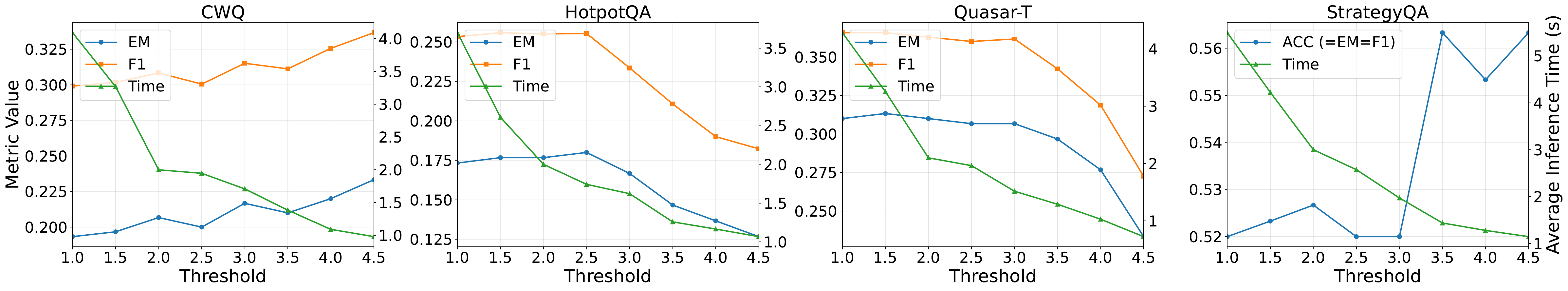}
    \caption{
        \textbf{Impact of triggering thresholds on effectiveness and inference speed.}
        We vary the TU threshold for expert activation. Larger thresholds reduce triggering frequency and improve inference speed, while downstream performance remains comparatively stable across a broad threshold range.
    }
    \label{fig:appendix_threshold_abl}
\end{figure*}

Figure~\ref{fig:appendix_threshold_abl} studies the effect of the TU triggering threshold. Increasing the threshold makes expert activation more selective, which reduces routing frequency and improves inference speed. At the same time, downstream metrics remain relatively stable across the tested range, indicating that DMoE does not require a narrowly tuned threshold to maintain performance. This behavior suggests that the TU criterion captures a useful region where expert activation is most needed, while allowing practitioners to adjust the threshold to trade off effectiveness and efficiency.

\subsection{Router: Impact of Retriever Choice}
\label{sec:appendix_retriever_choice}

\begin{table*}[t]
  \caption{\textbf{Impact of retriever choice for routing.} We compare the BM25 router used in the main experiments with SGPT-based dense retrieval. SGPT improves StrategyQA but does not consistently improve the other benchmarks, while using more GPU memory.}
  \label{tab:appendix_retriever_choice}
  \begin{center}
    \begin{small}
      \begin{sc}
        \setlength{\tabcolsep}{8pt}
        \renewcommand{\arraystretch}{1.12}
        \begin{tabular}{llcccc}
          \toprule
          \textbf{Dataset} & \textbf{Router} & \textbf{EM} & \textbf{F1} & \textbf{Avg. Time (s)} & \textbf{Avg. GPU (GB)} \\
          \midrule
          HotpotQA  & BM25 & \textbf{0.1800} & 0.2553 & 5.65 & \textbf{3.05} \\
          HotpotQA  & SGPT & \textbf{0.1800} & \textbf{0.2603} & \textbf{5.28} & 3.65 \\
          \midrule
          StrategyQA & BM25 & 0.5667 & 0.5667 & 8.51 & \textbf{2.62} \\
          StrategyQA & SGPT & \textbf{0.6067} & \textbf{0.6067} & \textbf{7.52} & 3.22 \\
          \midrule
          CWQ       & BM25 & \textbf{0.2467} & \textbf{0.3479} & 5.57 & \textbf{2.75} \\
          CWQ       & SGPT & 0.2267 & 0.3328 & \textbf{4.95} & 3.36 \\
          \midrule
          Quasar-T  & BM25 & \textbf{0.3133} & \textbf{0.3658} & 6.18 & \textbf{2.91} \\
          Quasar-T  & SGPT & 0.2933 & 0.3458 & \textbf{5.97} & 3.51 \\
          \bottomrule
        \end{tabular}
      \end{sc}
    \end{small}
  \end{center}
  \vskip -0.1in
\end{table*}

Table~\ref{tab:appendix_retriever_choice} studies whether replacing the default BM25 lexical router with SGPT dense retrieval improves expert selection. SGPT yields higher accuracy on StrategyQA and slightly higher F1 on HotpotQA, but it underperforms BM25 on CWQ and Quasar-T. The efficiency profile is also mixed: SGPT is modestly faster in these runs, but consistently consumes about $0.6$\,GB more GPU memory because dense retrieval introduces additional neural embedding computation and storage. Since SGPT does not provide a stable effectiveness gain and increases memory usage, we use BM25 as the default router in the main experiments.

\subsection{Experts: Impact of LoRA Rank and Scaling Factor}
\label{sec:appendix_lora_rank_alpha}

\begin{table}[t]
  \caption{\textbf{Impact of LoRA rank and scaling factor.} We vary the expert LoRA configuration $(r,\alpha)$ while keeping the remaining setup fixed. Higher rank or larger scaling does not yield consistent gains across benchmarks.}
  \label{tab:appendix_lora_rank_alpha}
  \begin{center}
    \begin{small}
      \begin{sc}
        \setlength{\tabcolsep}{5pt}
        \renewcommand{\arraystretch}{1.15}
        \resizebox{\columnwidth}{!}{%
        \begin{tabular}{lcccc}
          \toprule
          \textbf{Dataset} & \textbf{(4,16)} & \textbf{(8,16)} & \textbf{(8,32)} & \textbf{(16,32)} \\
          \midrule
          HotpotQA (EM)    & \textbf{0.1800} & 0.1700 & 0.1767 & 0.1733 \\
          CWQ (EM)         & \textbf{0.2467} & 0.2400 & 0.2300 & 0.2433 \\
          Quasar-T (EM)    & \textbf{0.3133} & 0.2933 & 0.2933 & 0.2900 \\
          StrategyQA (Acc.) & 0.5667 & 0.5767 & \textbf{0.5800} & 0.5767 \\
          \bottomrule
        \end{tabular}%
        }
      \end{sc}
    \end{small}
  \end{center}
  \vskip -0.1in
\end{table}

Table~\ref{tab:appendix_lora_rank_alpha} evaluates the sensitivity of DMoE experts to LoRA rank $r$ and scaling factor $\alpha$. The default configuration $(4,16)$ achieves the best result on HotpotQA, CWQ, and Quasar-T, while StrategyQA slightly favors $(8,32)$. Overall, increasing rank or scaling factor does not produce a consistent improvement, suggesting that the expert adapters do not require large capacity to capture the targeted knowledge units used in our setting. We therefore use $(4,16)$ as the default configuration because it provides the strongest overall trade-off between effectiveness and parameter efficiency.

\section{Scalability and Cost Analysis}
\label{sec:appendix_scalability_cost}

\subsection{Expert Count and Storage Footprint}

In DMoE, the number of experts is not a fixed architectural hyperparameter as in standard MoE models. It is determined by how external knowledge is partitioned into knowledge units. In our experiments, we instantiate one expert for each passage-level knowledge unit, resulting in 27{,}613 corpus-aligned experts for the Llama-3.2-1B expert bank. This design makes the expert bank naturally aligned with the retrieval corpus, but it is not the only possible granularity: in deployments with stronger storage constraints, multiple passages can be grouped into a document-level or cluster-level expert; conversely, finer units can be used when routing precision is more important.

Each expert is lightweight. With the default LoRA configuration, one expert contains 122{,}880 trainable parameters, corresponding to approximately 481\,KiB on disk. The full 27{,}613-expert bank occupies about 13.08\,GiB of disk storage. Importantly, this bank size does not translate into GPU memory pressure during inference. DMoE keeps the expert bank on disk and loads only the top-$k$ selected experts at triggered decoding steps; all inactive experts remain outside GPU memory. Therefore, increasing the expert bank mainly affects disk footprint and index management, while per-step GPU memory depends on the small number of active experts rather than the total number of stored experts.

\subsection{Training and Update Cost}

DMoE also differs from SFT-LoRA in how updates are performed. SFT-LoRA trains a single adapter on a merged corpus-scale dataset, so adding or modifying knowledge typically requires re-merging the training data and retraining or continuing training a shared adapter. In contrast, DMoE trains one small LoRA expert per knowledge unit independently. Existing expert checkpoints are skipped automatically, so when the corpus grows, only experts for newly added knowledge units need to be trained.

In our current implementation, training one new passage-level expert takes about 10 seconds on a single NVIDIA A100 GPU. Training is embarrassingly parallel because experts are independent: the 27{,}613-expert bank corresponds to roughly 76.7 A100 GPU-hours under this per-expert timing, and wall-clock time can be reduced by distributing experts across multiple GPUs. Updating obsolete knowledge is similarly local. One can delete the corresponding adapter directory and remove the document from the BM25 index, without retraining the backbone or unrelated experts.

\subsection{Scalability Trade-off}

The main scalability trade-off is therefore not the number of active parameters at inference time, but the granularity of the expert bank. A larger bank provides more fine-grained routing targets but increases disk usage and index size; a smaller bank reduces storage and management overhead but may merge heterogeneous knowledge into the same expert. This trade-off is controllable because knowledge units can be defined at different granularities before expert construction.

We consider two related but distinct dimensions of expert capacity. First, the capacity of each individual expert is controlled by the LoRA rank and scaling factor; as shown in Appendix~\ref{sec:appendix_lora_rank_alpha}, this effect is non-monotonic, and the default $(r=4,\alpha=16)$ setting is competitive or best on three of the four datasets. Second, the total expert-bank size is controlled by how many knowledge units are represented as separate experts. Table~\ref{tab:appendix_expert_bank_scaling} reports the latter ablation.

\begin{table}[t]
  \caption{\textbf{Expert-bank scaling ablation.} We reduce the expert bank to different fractions of the full corpus-aligned bank and report downstream performance. The full bank is not uniformly best, indicating that DMoE is not brittle to the exact expert count.}
  \label{tab:appendix_expert_bank_scaling}
  \begin{center}
    \begin{small}
      \begin{sc}
        \setlength{\tabcolsep}{4pt}
        \renewcommand{\arraystretch}{1.15}
        \resizebox{\columnwidth}{!}{%
        \begin{tabular}{lccccc}
          \toprule
          \textbf{Dataset} & \textbf{Full} & \textbf{1/2} & \textbf{1/5} & \textbf{1/10} & \textbf{1/100} \\
          \midrule
          HotpotQA  & 0.1800 & 0.1800 & 0.1800 & \textbf{0.1833} & 0.1767 \\
          StrategyQA & 0.5667 & 0.5800 & \textbf{0.5833} & 0.5700 & 0.5767 \\
          CWQ       & \textbf{0.2467} & 0.2367 & 0.2433 & 0.2300 & 0.2300 \\
          Quasar-T  & \textbf{0.3133} & \textbf{0.3133} & 0.3033 & 0.3067 & 0.3100 \\
          \bottomrule
        \end{tabular}%
        }
      \end{sc}
    \end{small}
  \end{center}
  \vskip -0.1in
\end{table}

The results show reasonably stable performance across substantial reductions in bank size. HotpotQA slightly improves at 1/10 of the bank, StrategyQA is best at 1/5, while CWQ and Quasar-T favor the full bank or tie with it. This pattern indicates that DMoE does not require a narrowly tuned expert count to remain functional.

\section{Prompt Design}
\label{sec:appendix_prompt_design}

This appendix lists the prompt templates used for document augmentation and evaluation. The augmentation prompts are used to construct document-specific parametric knowledge representations, while the evaluation prompt follows the few-shot reasoning format used in our experiments.

\newtcolorbox{promptbox}[1]{
  enhanced,
  width=0.92\linewidth,
  center,
  colback=black!2,
  colframe=black!45,
  colbacktitle=black!60,
  coltitle=white,
  fonttitle=\bfseries\small,
  title={#1},
  arc=2mm,
  boxrule=0.8pt,
  left=1.2em,
  right=1.2em,
  top=0.8em,
  bottom=0.8em,
  boxsep=0pt,
  titlerule=0pt,
  toptitle=0.45em,
  bottomtitle=0.45em,
}

\subsection{Prompts for Document Augmentation}

To construct document-specific parametric knowledge representations, we employ two augmentation techniques: multi-style rewriting and QA generation.

\paragraph{Document Rewriting}
We utilize rewriting to diversify surface forms while preserving semantic content.

\begin{promptbox}{Prompt 1: Document Rewriting}
\small
Rewrite the following passage.

While keeping the entities, proper nouns, and key details such as names, locations, and terminology intact, create a new version of the text that expresses the same ideas in a different way. Make sure the revised passage is distinct from the original one, but preserves the core meaning and relevant information.

Passage: passage
\end{promptbox}

\paragraph{QA Pair Generation}
This prompt is used to generate multiple QA pairs from a passage for downstream training.

\begin{promptbox}{Prompt 2: Question-Answer Generation}
\small
I will provide a passage of text, and you need to generate three different questions based on the content of this passage. Each question should be answerable using the information provided in the passage. Additionally, please provide an appropriate answer for each question derived from the passage.

You need to generate the question and answer in the following format:

[

\{

"question": "What is the capital of France?",

"answer": "Paris",

"full\_answer": "The capital of France is Paris."

\},

]

This list should have at least three elements. You only need to output this list in the above format.

Passage: passage
\end{promptbox}

\subsection{Prompt for HotpotQA}

Following prior work such as FLARE\citep{jiang2023active} and DRAGIN\citep{su2024dragin}, we adopt a similar few-shot prompting strategy. All compared methods use the same prompt format for a fair evaluation.

We use reasoning exemplars to guide the model toward concise multi-hop answers.

\begin{promptbox}{Prompt 3: Few-Shot Prompt for HotpotQA}
\small
You should reference the knowledge provided below and combine it with your own knowledge to answer the question. Please follow the format of the example I provided above.

Question: Jeremy Theobald and Christopher Nolan share what profession? Answer: Jeremy Theobald is an actor and producer. Christopher Nolan is a director, producer, and screenwriter. Therefore, they both share the profession of being a producer. So the answer is producer.

Question: Were Lonny and Allure both founded in the 1990s? Answer: Lonny (magazine) was founded in 2009. Allure (magazine) was founded in 1991. Thus, of the two, only Allure was founded in 1990s. So the answer is no.

... [additional examples truncated for brevity] ...

Question: Were Scott Derrickson and Ed Wood of the same nationality?
\end{promptbox}

\end{document}